\documentclass[runningheads]{llncs}

\usepackage{graphicx}
\usepackage{cite}
\usepackage{url}
\usepackage{verbatim}
\usepackage{amsmath,amssymb,amsfonts}
\usepackage{algorithmic}
\usepackage{mathtools}
\DeclarePairedDelimiter{\ceil}{\lceil}{\rceil}
\usepackage{graphicx}
\usepackage{textcomp}
\usepackage{multirow}
\usepackage[shortlabels]{enumitem}
\usepackage{tikz}
\usepackage[misc,geometry]{ifsym}
\usepackage[bottom]{footmisc}
\usepackage{subcaption}
\usepackage{hyperref}

\begin{document}
\title{Improving Action Quality Assessment Using Weighted Aggregation}
%
%
\author{Shafkat Farabi(\Letter)\inst{1}\orcidID{0000-0003-4712-1208} \and
Hasibul Himel\inst{1}\orcidID{0000-0002-9246-2477} \and
Fakhruddin Gazzali\inst{1}\orcidID{0000-0001-5609-5852} \and
Md. Bakhtiar Hasan\inst{1}\orcidID{0000-0001-8093-5006} \and 
Md. Hasanul Kabir\inst{1}\orcidID{0000-0002-6853-8785} \and 
Moshiur Farazi\inst{2}\orcidID{0000-0003-1494-5921}
}
\authorrunning{Farabi et al.}
%
\institute{Islamic University of Technology, Gazipur, Bangladesh\\
\email{\{shafkatrahman, hasibulhaque, fakhruddingazzali, bakhtiarhasan, hasanul\}@iut-dhaka.edu}\\
 \and
Data61-CSIRO, Canberra, Australia\\
\email{moshiur.farazi@data61.csiro.au}}
\maketitle              
\begin{abstract}
Action quality assessment (AQA) aims at automatically judging human action based on a video of the said action and assigning a performance score to it. The majority of works in the existing literature on AQA divide RGB videos into short clips, transform these clips to higher-level representations using Convolutional 3D (C3D) networks, and aggregate them through averaging. These higher-level representations are used to perform AQA. We find that the current clip level feature aggregation technique of averaging is insufficient to capture the relative importance of clip level features. In this work, we propose a learning-based weighted-averaging technique. Using this technique, better performance can be obtained without sacrificing too much computational resources. We call this technique Weight-Decider(WD). We also experiment with ResNets for learning better representations for action quality assessment.  We assess the effects of the depth and input clip size of the convolutional neural network on the quality of action score predictions. We achieve a new state-of-the-art Spearman's rank correlation of 0.9315 (an increase of 0.45\%) on the MTL-AQA dataset using a 34 layer (2+1)D ResNet with the capability of processing 32 frame clips, with WD aggregation.
\end{abstract}
\section{Introduction}
Action quality assessment (AQA) addresses the problem of developing a system that can automatically judge the quality of an action performed by a human. This is done by processing a video of the performance and assigning a score to it. The motivation to develop such a system stems from its potential use in applications such as health care  \cite{measureingQualityOfExcercise},  sports video analysis \cite{WhatandHow}, skill discrimination for a specific task \cite{aqamultipleaction}, assessing the skill of trainees in professions such as surgery \cite{surgiacalSkill}, etc. 

\begin{figure}[t]
 \center
\includegraphics[width=\textwidth]{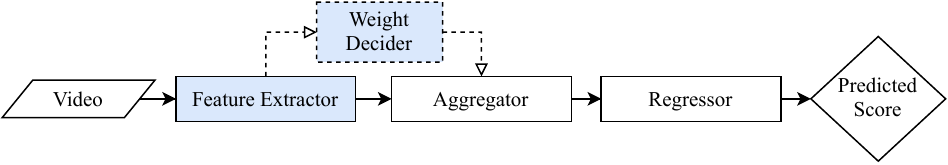}

  \caption{\textbf{Overview of a general AQA pipeline and our improvement over it}. Generally, the input video is divided into clips. A feature extractor extracts features from these clips. These features are then aggregated into a video level feature vector. A linear-regressor predicts action quality scores based on this feature vector. We introduce a Weight-Decider module to this architecture, which proposes weights based on the clip level features for better aggregation. Additionally, we use a ResNet instead of the commonly used C3D.}
  

  \label{img:general}

\end{figure}

Almost all existing works have treated AQA as a regression problem \cite{aqamultipleaction, asessQualityOfActions, learningToScoreOlympic}. As shown in Figure \ref{img:general}, most approaches boil down to dividing an RGB video of the action in multiple clips, extracting higher-level features from each clip, aggregating them, and then training a linear-regressor to predict a score based on these aggregated features. Most of these works \cite{asessQualityOfActions, learningToScoreOlympic, s3dAQA} utilize a convolutional neural network \cite{CNNoriginal} to extract complex higher-level features and simple averaging to aggregate the features. The best performing models \cite{WhatandHow, asessQualityOfActions, learningToScoreOlympic} make use of the Convolutional-3D (C3D) network \cite{c3dPaper} with average aggregation. 

The majority of works in AQA \cite{asessQualityOfActions, learningToScoreOlympic, s3dAQA,WhatandHow} aggregate clip level features into a video level feature vector by simply averaging them. We think this fails to preserve temporal information present in the data. We hypothesize that a more sophisticated method would improve the overall performance of the pipeline. We propose one such technique by introducing a module called Weight-Decider(WD). This module inspects the feature vectors extracted from individual clips and proposes a corresponding weight vector. Finally, these weight vectors can be used to calculate a weighted average of the clip-level feature vectors. In this way, the final video level feature vector contains more contributions from the important features of each clip. This is similar to how real-world judges base their final scoring on key mistakes/skills of the performer and not on an average of all the moments in the action. We design the WD as a shallow neural network that can be trained along with the rest of the AQA pipeline. In our experiments, we show the performance of the AQA pipeline to improve when using WD as aggregation. 

Spatio-temporal versions of ResNets capable of processing videos have been proposed by \cite{3DresnetPaper} and \cite{2p1dPaper}. These approaches have achieved state-of-the-art results in the task of action recognition and outperform the C3D network \cite{3DresnetPaper}. Hence, we decide to use spatio-temporal ResNets as feature extractors for performing AQA. We experiment with various 3D and (2+1)D ResNet feature extractors on the MTL-AQA dataset \cite{WhatandHow}. We find that 3D and (2+1)D ResNets of depth 34 and 50 with pretraining on large-scale action recognition datasets have performance comparable to the state-of-the-art. We see that (2+1)D and 3D convolutions perform fairly similarly. For 34 layer (2+1)D ResNets, we experiment with 3 different versions that can process 8, 16, or 32 frame clips at once and find the 32 frame clip version to clearly outperform the rest. Our results suggest that processing longer clips is more beneficial than going deeper with convolutions. The 34 layers (2+1)D ResNet with WD processing 32 frame clips achieves a Spearman's rank correlation of 0.9315 on the MTL-AQA dataset, achieving a new state-of-the-art.

\textbf{Contributions:}
\begin{itemize}  
    \item We propose a novel learning-based light-weight aggregation technique called Weight-Decider and demonstrate that it can improve the AQA pipeline's performance.
     \item To the best of our knowledge, this is the first work to do a comparative analysis of the effect of the depth, convolution type, and input clip size of the ResNet feature extractor on the final quality of the scores predicted in AQA.
    \item One of our approaches outperforms all the previous works in AQA on the MTL-AQA dataset.
\end{itemize}

\section{Related Work}
Pirsiavash et al.\cite{asessQualityOfActions} proposed a novel dataset containing videos of Diving and Figure-skating annotated with action quality scores by expert human judges. They provided Discrete Cosine Transform (DCT) and extracted human pose as inputs to a Support Vector Regressor, which predicted the score. 

More recent works have utilized the Convolutional 3D (C3D) network \cite{c3dPaper} as a feature extractor. Parmar and Morris \cite{learningToScoreOlympic} proposed three architectures, C3D-SVR, C3D-LSTM, C3D-LSTM-SVR, all of which used features extracted from short video clips using C3D network, and later aggregated them and predicted an action score using Support Vector Regressor (SVR) and Long Short-Term Memory (LSTM).In a later work, Parmar and Morris\cite{WhatandHow} took a multitask approach towards action quality assessment. They released a novel AQA dataset called MTL-AQA and proposed a multi-task learning-based C3D-AVG-MTL framework that extracted features using the C3D network and aggregated these through averaging. They trained these features to do well in score prediction, action classification, and generating captions. Tang et al.\cite{uncertainAQA} took a probabilistic approach (MUSDL). They divided 103 frame videos into 10 overlapping 16 frame clips, used I3D\cite{i3dPaper} architecture to extract clip level features, averaged as aggregation, and finally predicted parameters of a probabilistic distribution from which the final score prediction was sampled. The authors calculated 7 different scores corresponding to 7 judges for Olympic scoring and summed up the 5 scores in the middle. With this advantage over simple regression-based methods which directly predict the score, this approach achieved a SOTA spearman's correlation of 0.9273. Diba et al.\cite{channelcorr} used a method called ``STC Block" for action recognition. This is similar to our proposed aggregation method. However, they utilize this on spatial and temporal features separately after each convolution layer for action recognition, whereas our method is applied to the output of the CNN to aggregate clip level spatiotemporal features for performing AQA.

Our proposed approach differs from these works in that we use 3D and (2+1)D ResNets as the feature extractor and we aggregate these features using the WD network, which is a lightweight and learning-based feature aggregation scheme.


\section{Our Approach}

\subsection{General Pipeline Overview}

Let, $V = \{F_{t}\}_{t=1}^{L}$ be the input video containing $L$ frames, where $F_{t}$ denotes the $t^{th}$ frame. It is divided into $N$ non-overlapping clips, each of size $n=\ceil{\frac{L}{N}}$. Thus we define the $i^{th}$ clip as $C_{i} = \{ F_{j} \}_{j=i\times n}^{i\times (n+1)-1}$. The feature extractor takes in a clip $C_{i}$ and outputs a feature vector $f_{i}$. For the feature extractor, we  experiment with 3D ResNets\cite{resnetPaper} and (2+1)D ResNets\cite{2p1dPaper} with varying depth and input clip size. Next, we aggregate these clip level features to obtain a global video level representation. Finally, a linear-regressor is trained to predict the score from the video level feature representation. Following the majority of previous works \cite{WhatandHow, aqamultipleaction, asessQualityOfActions, learningToScoreOlympic}, we model the problem as linear regression. This makes sense as the action quality score is a real number. 
To experiment with the relation of the ResNet feature extractor's depth with the AQA pipeline's ability to learn, we experiment with 3 different depths:
\begin{itemize}
    \item \textbf{34 layer}: We experiment with both 34 layer 3D and (2+1)D ResNets. The only difference being 3D ResNets use $3\times3\times3$ convolution kernels, on the other hand (2+1)D ResNets use a $1\times3\times3$ convolution followed by $3\times1\times1$ convolution\cite{2p1dPaper}. We take the final average-pool layer output (512 dimensional) and pass it through 2 fully-connected layers having $256$ and $128$ units. The 3D ResNet takes input 16 frame clips. On the other hand, we experiment with 3 different variations of (2+1)D 34-layer ResNet, processing 8, 16, and 32 frame clips using available pre-trained weights,

    \item \textbf{50 layer}: We experiment with  50 layer 3D and (2+1)D ResNets. The final average-pool layer outputs a feature vector of size $2048$. We take this feature vector and input it into 3 fully-connected layers having $512$, $256$, and $128$ units. Only 16 frame clips are processed.
    \item \textbf{101 layer}: We experiment with 101 layer 3D ResNet. The remaining details about the input clip size and output feature vector processing are identical to the 50 layer ResNets.

\end{itemize}

\subsection{Feature Aggregation}
Most of the previous works dealing with AQA process the entire input video by first dividing it into multiple smaller clips of equal size, due to memory and computational budget. Most CNNs are designed to process 8, 16, or 32 frames at once. Then the features extracted by the CNN are aggregated to form a video level feature description. Next, a linear-regressor predicts the final score based on this feature description.
\begin{figure}[t]
\center

 \includegraphics[width=\textwidth]{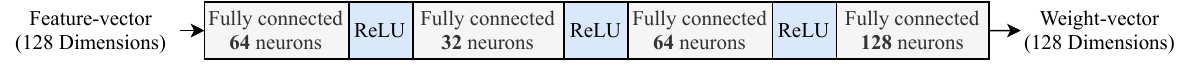}

  \caption{The architecture of the WD network.}

  \label{img:weight-decider}

\end{figure}
The best performing works aggregated the clips by simply averaging them \cite{WhatandHow, uncertainAQA, learningToScoreOlympic}. Some other works \cite{learningToScoreOlympic, aqamultipleaction} aggregated using LSTMs\cite{lstm}. However, LSTM networks, although make sense in theory because of their ability to handle time sequences, perform worse due to the lack of big-enough datasets dedicated to AQA.

We propose that simply averaging the clip-wise features is an ineffective measure. It should not be able to preserve the temporal information available in the data. This follows from the fact changing the order of the clip level features will generate the same average and hence the same score prediction. Furthermore, expert human judges focus more on mistakes and deviations of the performers and these have a bigger impact on the score. Hence we think, a weighted averaging technique might be more suitable, as the linear-regressor will be able to base its decision on features more important from each clip.

More concretely, if the feature vector extracted from clip $C_{i}$ is $f_{i}$, we propose the video level feature vector as
\begin{equation}
\label{eq1}
    f_{video} = \sum_{i=1}^{N} ( f_{i} \odot w_{i} )
\end{equation}
where $w_{i}$ is a weight vector corresponding to the feature vector $f_{i}$ and $\odot$ represents Hadamard Product or elementwise multiplication.  $w_{i}$ is of the same dimensions as $f_{i}$ and learned using a small neural network of 4 layers. This smaller neural network takes as input 128-dimensional feature vector $f_{i}$ and runs it through fully connected layers containing 64, 32, 64, and 128 neurons. All but the final layers employ a ReLU activation function. The architecture is explained in Figure \ref{img:weight-decider}. Finally, to ensure the weights corresponding to the same element of different weight vectors sum up to one, a softmax is applied along with the corresponding elements of all the weight vectors.
We call this shallow neural network Weight-Decider(WD).

\begin{equation}
\label{eq2}
    w_{i}^{'} = WD(f_{i})
\end{equation}
\begin{equation}
\label{eq3}
    [w_{0} \quad w_{1} \quad \dots \quad w_{N}] = softmax( [w_{0}^{'} \quad w_{1}^{'} \quad \dots \quad w_{N}^{'}])
\end{equation}

Finally, the linear-regressor can predict the score using the feature vector $f_{video}$ as proposed by equation \ref{eq1}. 

\begin{figure}[t]
   \includegraphics[width=\textwidth]{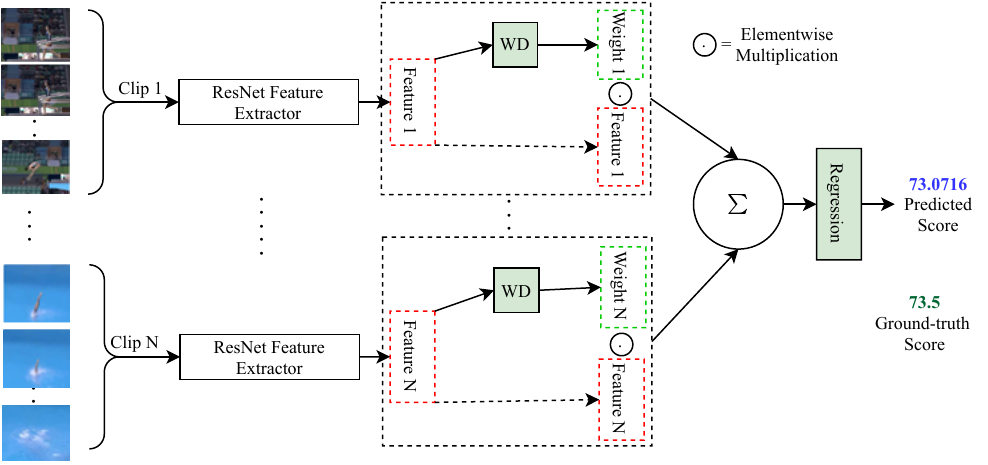}
    
  
  \caption{The video is divided into clips and each clip is processed by a ResNet feature extractor. The ResNet can have depths 34, 50, or 101. Elements with a solid border indicate trainable modules. The solid bordered white modules are initialized with pretrained weights. The solid bordered green modules are trained from scratch. Best viewed in color.}
  \label{img:final}
\end{figure}

The proposed WD module can be used with any of the feature extractors we planned to use in our experiments. WD replaces averaging as aggregation within the typical AQA pipeline. It can be trained with the rest of the AQA model in an end-to-end manner.  

Adding WD to the AQA pipeline to replace averaging as aggregation does not cost much in terms of computational resources. To see this, recall that the WD module has 3 hidden layers and an output layer, consisting of 64, 32, 64, and 128 neurons in that order. Each of these layers contains a weight matrix and a bias.  The first hidden layer takes 128 inputs and has 64 neurons. Thus the weight matrix is of dimensions $64 \times 128$. Hence, number of trainable parameters in this layer (including the bias term) is $(64 \times 128 + 1) = 8193$. Similarly, the second hidden layer, third hidden layer, and the output layers contain $2049$, $2049$, and $8193$ trainable parameters correspondingly. By summing up all the trainable parameters in each layer, we can see that WD only contains $20484$ trainable parameters. Spatio-temporal ResNet feature extractors have millions of trainable parameters. Hence, additional resources required to train the WD module used in an AQA pipeline are not much. As an example, the 3D and (2+1)D ResNet-34 feature extractors we are using contain  63.6 million trainable parameters \cite{param}. Thus, using the WD module on top of ResNet-34 would increase the number of trainable parameters by approximately $0.03\%$. The ratio would be smaller for a deeper ResNet. Hence, our proposed WD does not require many computational resources to be incorporated into the pipeline. Our experiments support this. We found the training time for 32-frame ResNet(2+1)D-34 to be 4557 seconds per epoch, and the inference time to be 604 seconds per epoch. The introduction of WD increases the training time per epoch by 24 seconds (0.52\%) and the inference time by 21 seconds (3.47\%). 
\section{Experiments}

\subsection{MTL-AQA Dataset\cite{WhatandHow}} The biggest dataset dedicated to AQA. It contains 1412 video samples split into 1059 training and  353 test samples. The samples are collected from 16 Olympic dive events. Each sample is 103 frames long and accompanied by the final action quality scores from expert Olympic judges. It was published by Parmer and Morris\cite{WhatandHow}. We used the exact same train/test split made public in their work\cite{WhatandHow}.

\subsection{Evaluation Metric:}

In line with previously published literature, we use Spearman's rank correlation as the evaluation metric. This metric measures the correlation between two sequences containing ordinal or numerical data. It is calculated using the equation: 
\begin{equation}
    \rho = 1 - \frac{6\sum d_{i}^{2}}{n(n^{2}-1)}
\end{equation}
$\rho$ = Spearman's rank correlation\\
$d_{i}$= The difference between the ranks of corresponding variables\\
$n$ = Number of observations

\subsection{Implementation Details:}
We implemented our proposed methods using PyTorch\cite{pytorch}. All the 3D ResNets and (2+1)D ResNets processing 16 frame clips were pre-trained on Kinetics-700\cite{Kinetics700}  dataset\footnote{Weights available at: \url{https://github.com/kenshohara/3D-ResNets-PyTorch}}. The (2+1)D Resnets processing 8 and 32 frame clips were pre-trained on IG-65M dataset\cite{ig65m} and fine tuned on Kinetics-400\cite{Kinetics700}  dataset\footnote{Weights available at: \url{https://github.com/moabitcoin/ig65m-pytorch}}.

For each ResNet, we separately experimented using both averaging and  WD as feature aggregation. We temporally augmented by randomly picking an ending frame from the last $6$ and chose the preceding $96$ frames for processing. The frames were resized to $171 \times 128 $ and center cropped to $112 \times 112 $. Random horizontal flipping was applied. Batch-normalization was used for regularization. We defined the loss function as a sum of L2 and L1 loss between the predicted score and ground-truth score as Parmar and Morris\cite{WhatandHow} suggested. We trained the network using the ADAM optimizer\cite{adam} for $50$ epochs. We used a learning rate of 0.0001 for modules with randomly initialized weights and  0.00001 for modules with pretrained weights. Train and Test batch sizes were 2 and 5.


\subsection{Results on MTL-AQA Dataset:}

\begin{table}[t]
\setlength\tabcolsep{0pt}
\caption{Performance comparison of the various types of ResNets as feature extractors and varying clip length in our pipeline}
\begin{subtable}{.57\textwidth}
\caption{Effect of various types of ResNets}
\label{tab:performance_resnets}
\begin{tabular*}{0.9\linewidth}{@{\extracolsep{\fill}} c c c c }

\hline
\multirow{2}{*}{\begin{tabular}[c]{@{}c@{}}\textbf{Depth}\end{tabular}} &
  \multirow{2}{*}{\begin{tabular}[c]{@{}c@{}}\textbf{ Convolution}\end{tabular}} &
  \multicolumn{2}{c}{\textbf{Aggregation}} \\ \cline{3-4} 
   &                    & \textbf{Average} & \textbf{WD} \\ \hline

\multirow{2}{*}{ResNet-34}  & 3D     & 0.8982 & 0.8951  \\ 
                     & (2+1)D & 0.8932 & 0.8990  \\ \hline
\multirow{2}{*}{ResNet-50}  & 3D     & 0.8880 & 0.8935  \\ 
                     & (2+1)D & 0.8818 & 0.8814  \\ \hline 
ResNet-101 & 3D     & 0.6663 & 0.6033 \\ \hline              
\end{tabular*}%
\end{subtable}
\begin{subtable}{.43\textwidth}
\centering
\caption{Effect of varying input clip size in ResNet(2+1)D-34}
\label{tab:clip_comparisom}
\begin{tabular*}{0.9\linewidth}{@{\extracolsep{\fill}} c c c }
\hline
\multirow{2}{*}{\begin{tabular}[c]{@{}c@{}}\textbf{Clip length}\\ \textbf{(Input Frames)}\end{tabular}} & \multicolumn{2}{c}{\textbf{Aggregation}} \\ \cline{2-3} 
   & \textbf{Average} & \textbf{WD} \\ \hline
8  & 0.8570  & 0.8853        \\ 
16 & 0.8932  & 0.8990        \\ 
32 & 0.9289  & 0.9315        \\ \hline
\end{tabular*}
\end{subtable}


\end{table}

\begin{figure}[h]
    \centering

  \includegraphics[width=\textwidth]{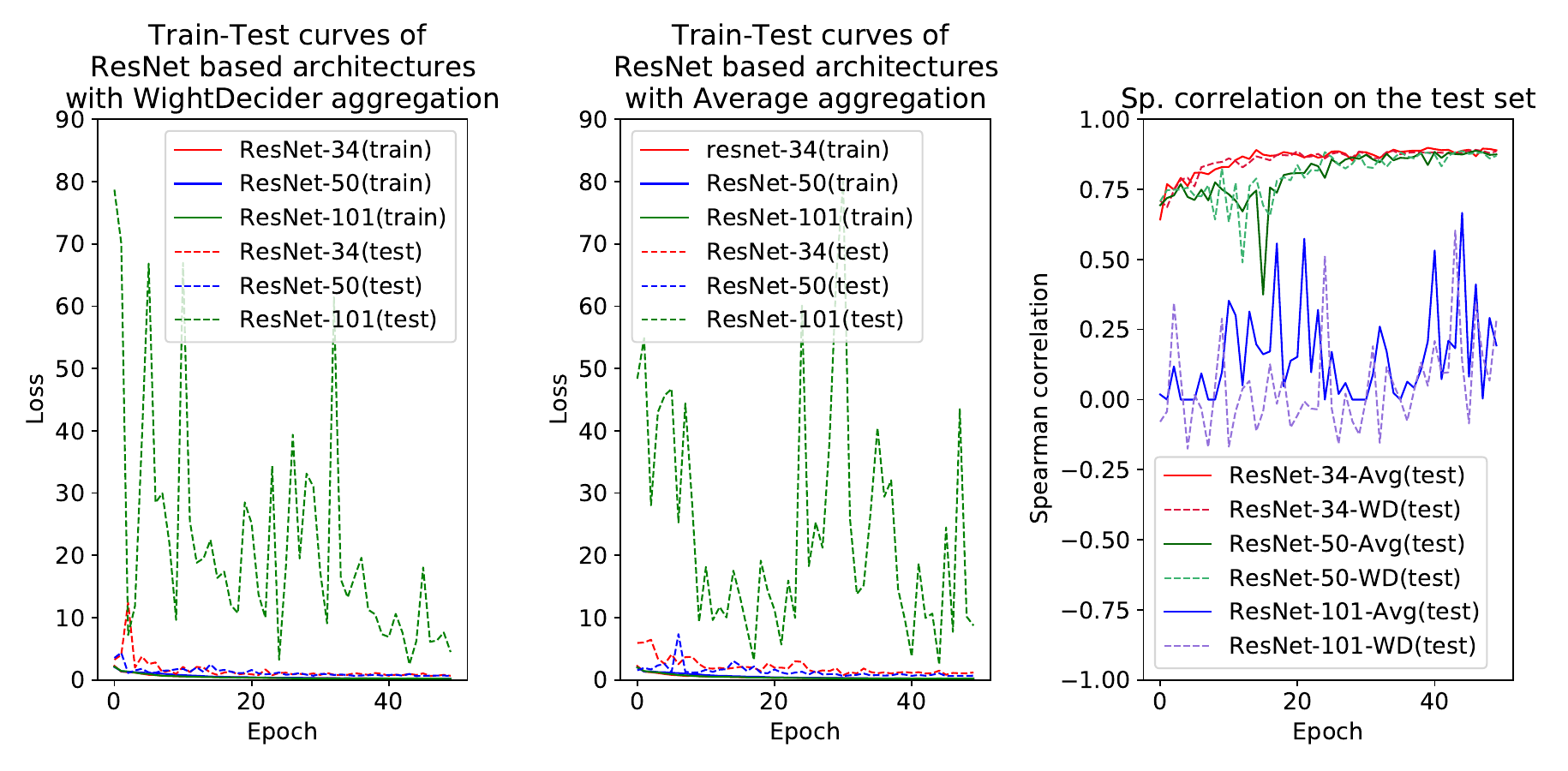}
   \caption{\textbf{Train and Test curves obtained from training the pipeline using 3D-ResNet 34, 50, and 101 as feature extractors.} Notice that ResNet-101 based architectures show signs of significant overfitting compared to ResNet-34 and ResNet-50 based architectures.}
    \label{img:train_test}
    
    
\end{figure}

In Table \ref{tab:performance_resnets}, we present the experiment results of varying the depth of the ResNet feature extractor and the aggregation scheme.  We can see that 34 layer (2+1)D ResNet with WD as aggregation performs the best with a Spearman's correlation of $0.8990$.  Increasing the depth to 50 layers somewhat decreases Spearman's correlation. However, the results are still competitive. At 101 layer depth, even when initialized with pretrained weights from Kinetics\cite{Kinetics700}, overfitting occurs fairly quickly. The overfitting is also evident from the train/test curves presented in Figure \ref{img:train_test}. The likely reason behind this is the increased number of parameters in the feature extractor. This leads us to establish that the current biggest AQA dataset has enough data to train a 34-layer and 50-layer ResNet feature extractor with generalization, however it overfits the 101-layer ResNet feature extractor.

 Because of how (2+1)D ResNets are designed, they have a similar parameter count to their 3D counterparts \cite{2p1dPaper}. Because the overfitting is occurring due to the high parameter count, we do not repeat the experiment with a (2+1)D ResNet-101 feature extractor.

\subsubsection{Effect of clip length:} We check the effect of clip length on the performance. We take the best performing model from Table \ref{tab:performance_resnets} (ResNet(2+1)D-34-WD) and input various clip sizes. We experiment with clip sizes of 8 frames, 16 frames, and 32 frames. We vary the aggregation method as well.

From Table \ref{tab:clip_comparisom} we can see that the performance of the pipeline increases with the number of frames in each clip. We hypothesize that longer clips allow the ResNet to look for bigger patterns in the temporal dimension, which in turn enables the feature descriptors to be more informative. This enables the linear-regressor to better discriminate between similar-looking examples with fine-grained action quality differences.  From Table \ref{tab:clip_comparisom}, notice that using WD over simple averaging as aggregation gives a boost in performance. However, this performance boost is quite significant in case of 8 frame clips. We believe the reason behind this improved performance is that increasing clip size reduces the number of clips being averaged. Whatever detrimental effect the averaging might have, it will be more prominent when the number of objects being averaged is larger, and less when this number is smaller. Furthermore, CNNs with longer input clips can look at more frames, this in effect increases their temporal horizon. It follows that the feature vectors extracted would have a better encoding of action patterns across time, to begin with. Thus they perform well enough even with averaging as aggregation. But using WD increases performance nevertheless.
    
For qualitative results, refer to Table \ref{tab:qualitative}.

\begin{table*}[h]
\centering

\caption{Qualitative results. Every $16^{th}$ frame processed starting from frame 0 is shown. Italic scores correspond to WD aggregation, plain text scores correspond to average aggregation. The 8, 16, and 32 correspond to input clip sizes. }

 \label{tab:qualitative}
\resizebox{0.93\textwidth}{!}
{%
\begin{tabular}{ c c c c c | c c | c c}
\hline
\multirow{4}{*}{\textbf{Input Frames}}  & \multicolumn{4}{c|}{\textbf{ResNet-34}} & \multicolumn{2}{c|}{\textbf{ResNet-50}} & \textbf{ResNet-101} &
 \\
 & \multicolumn{4}{c|}{\textbf{Prediction}} &  \multicolumn{2}{c|}{\textbf{Prediction}} & \textbf{Prediction} & \textbf{True}
\\\cline{2-8}
 & \multicolumn{3}{c}{\textbf{(2+1)D}} & \multirow{2}{*}{\textbf{3D}} & \multirow{2}{*}{\textbf{(2+1)D}} & \multirow{2}{*}{\textbf{3D}} & \multirow{2}{*}{\textbf{3D}} & \textbf{Score}\\\cline{2-4}
 & \textbf{8} & \textbf{16} & \textbf{32} & & & & & \\\hline
\multirow{4}{*}{\includegraphics[scale=1]{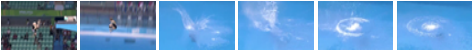}} &  &  & &  &  &  &  & \multirow{4}{*}{25.64} \\
& 54.84 & 30.46 & 8.39 & 7.29 & 33.23 & 34.10 & 45.22 & \\
 & {\textit{38.76}} & {\textit{18.11}} & {\textit{16.41}} & {\textit{22.93}} & {\textit{38.29}} & {\textit{29.93}} & {\textit{52.21}} & \\
 & & & & & & & & \\\hline
\multirow{4}{*}{\includegraphics[scale=1]{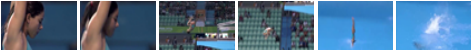}} &  &  & &  &  &  &  & \multirow{4}{*}{52.79} \\
& 66.94 & 59.69 & 47.92 & 67.92 & 43.57 & 58.30 & 122.20 & \\
 & {\textit{63.85}} & {\textit{40.88}} & {\textit{53.21}} & {\textit{63.62}} & {\textit{52.80}} & {\textit{52.31}} & {\textit{76.64}} & \\
 & & & & & & & & \\\hline
 \multirow{4}{*}{\includegraphics[scale=1]{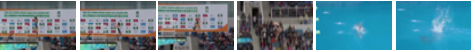}} &  &  & &  &  &  &  & \multirow{4}{*}{69.59} \\
& 71.46 & 71.34 & 69.39 & 83.34 & 67.38 & 80.41 & 167.60 & \\
 & {\textit{69.85}} & {\textit{64.90}} & {\textit{70.40}} & {\textit{81.31}} & {\textit{67.53}} & {\textit{74.25}} & {\textit{132.50}} & \\
 & & & & & & & & \\\hline
  \multirow{4}{*}{\includegraphics[scale=1]{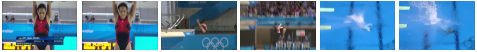}} &  &  & &  &  &  &  & \multirow{4}{*}{46.20} \\
& 67.13 & 46.16 & 27.73 & 34.25 & 44.06 & 46.61 & 54.28 & \\
 & {\textit{64.54}} & {\textit{32.29}} & {\textit{42.87}} & {\textit{39.62}} & {\textit{49.03}} & {\textit{47.13}} & {\textit{51.51}} & \\
 & & & & & & & & \\\hline
\end{tabular}
}
\end{table*}
\newpage

 The ground truth scores provided in the MTL-AQA \cite{WhatandHow} are taken from expert Olympic judges during a live broadcast of events. 7 judges independently score the athlete's performance on a scale of 0 (completely failed) to 10 (excellent). The 3 median scores are then added together and multiplied with a pre-determined difficulty degree to obtain the final score. This final score is the one provided in the ``True Score'' column of Table \ref{tab:qualitative}. Our various pipelines attempt to predict this final score from the input performance video.

\subsubsection{Comparison with the state of the art:}
In Table \ref{tab:compare_SOTA}, we compare our best performing models of each depth with previous works on the MTL-AQA dataset. For comparison, we combined C3D architecture with WD to test the result. For the C3D-WD model, 16 frame clips were used. The C3D portion of the model was initialized with Sports-1M\cite{sports1M} pretrained weights. We include this result in Table \ref{tab:compare_SOTA} as well. We can see that our (2+1)D ResNet-34 (32 frame)- WD outperforms all previous works in the literature. This shows the effectiveness of our approach. We can further see that 3D ResNet-50 (16 frame) obtains results comparable to the SOTA.  However, the ResNet-101 based approach overfits the dataset and hence performs poorly.

\begin{table}[t]
\centering

\caption{Comparison with the state of the art on the MTL-AQA dataset}
\label{tab:compare_SOTA}
\resizebox{0.55\textwidth}{!}
{
\begin{tabular}{ l c}
\hline
\textbf{Method}                                           & \textbf{Sp. Corr.}                   \\ \hline
Pose+DCT\cite{asessQualityOfActions}             &  0.2682                      \\
C3D-SVR\cite{learningToScoreOlympic}             &  0.7716                      \\
C3D-LSTM\cite{learningToScoreOlympic}            &  0.8489                      \\
MSCADC-STL\cite{WhatandHow}                      &  0.8472                      \\
MSCADC-MTL\cite{WhatandHow}                      &  0.8612                      \\
USDL-Regression\cite{uncertainAQA}               &  0.8905                      \\

C3D-AVG-STL\cite{WhatandHow}                     &  0.8960                      \\

C3D-AVG-MTL\cite{WhatandHow}                     &  0.9044                      \\
MUSDL \cite{uncertainAQA}                        &  0.9273                      \\ 
\hline
Ours C3D-WD                                           &  0.9223                      \\
\begin{tabular}[c]{@{}c@{}}Ours ResNet34-(2+1)D-WD (32 frame) \end{tabular} & \textbf{0.9315} \\ 

\begin{tabular}[c]{@{}c@{}}Ours ResNet50-3D-WD (16 frame) \end{tabular} & 0.8935 \\
\begin{tabular}[c]{@{}c@{}}Ours ResNet101-3D-AVG (16 frame) \end{tabular} & 0.6633 \\ 
\hline
\end{tabular}%
}
\end{table}

\section{Conclusion}
In this work, we proposed a ResNet-based regression-oriented pipeline for action quality assessment. We demonstrated experimentally that the MTL-AQA dataset has enough data to train 34 and 50 layer ResNet-based pipelines when initialized with pretrained weights from a related task (like action recognition). Our experiments suggest processing longer clips is more effective than using deeper ResNets. We also propose a lightweight learning-based aggregation technique called WD to replace simple averaging. Experiments show our methods to be more effective than previous works. In the future, we want to investigate with CNNs that can process longer clips (64 or higher) to see if this translates to better performance.

%
%

%
%
%
\bibliographystyle{splncs04}
\bibliography{mybibliography}

\end{document}